\documentclass{article} %
\usepackage{iclr2027_conference,times}

\usepackage{amsmath,amsfonts,bm}

\def\eqref#1{equation~\ref{#1}}

\def\1{\bm{1}}

\DeclareMathAlphabet{\mathsfit}{\encodingdefault}{\sfdefault}{m}{sl}
\SetMathAlphabet{\mathsfit}{bold}{\encodingdefault}{\sfdefault}{bx}{n}

\usepackage{hyperref}
\usepackage{url}
\usepackage{graphicx}
\usepackage{float}       %
\usepackage{amsmath}     %
\usepackage{amssymb}     %
\usepackage{booktabs}    %
\usepackage{multirow}    %
\usepackage{wrapfig}                            %
\usepackage[table]{xcolor}                      %
\definecolor{oursblue}{HTML}{2A78D6}            %
\graphicspath{{figs/}}
\usepackage[skip=4pt,font=small]{caption}   %

\title{MontageFlow: Montage-Consistent High Fidelity Bipolar EEG Generation}

\author{Sudarshan Regmi, Nabaraj Subedi, Arvind Pillai, Yu Yvonne Wu, Yuliang Chen, \\
\textbf{Michael V.~Heinz, Lisa Marsch, Nicholas C.~Jacobson, Andrew Campbell} \\
Dartmouth College \\
Correspondence: \href{mailto:sudarshan.regmi.gr@dartmouth.edu}{\texttt{sudarshan.regmi.gr@dartmouth.edu}}
}

\newcommand{\ours}{MontageFlow}

\newif\ifpreprint
\preprinttrue
\ifpreprint\iclrfinalcopy\fi
\begin{document}

\maketitle
\ifpreprint
  \lhead{Preprint}
  {\let\thefootnote\relax\footnotetext{Code is available at \url{https://github.com/sudarshanregmi/montageflow}.}}
\fi

\begin{abstract}
Generative modeling offers a promising solution to the scarcity of labeled clinical electroencephalography (EEG) data under strict privacy constraints. Although recent generative models report substantial improvements in generation fidelity, we show that these gains conceal a failure in the bipolar montage. In a bipolar montage, each channel is the voltage difference between two scalp electrodes. The bipolar channels therefore satisfy exact linear identities, which we term \textit{montage consistency}. Current bipolar EEG generators synthesize montage-inconsistent signals. We propose \textbf{MontageFlow}, a flow-matching model that enforces these identities by restricting its output to signals that satisfy them. We find that ensuring validity improves fidelity too. Furthermore, to retain the EEG characteristics that oversmoothing washes out, we condition the model on a physiological state vector encoded from the whole EEG window. For generation, this vector is drawn from a prior fitted in latent space. Across seven datasets in the bipolar montage, our model improves TS-FID by 1.3–5.7× over the strongest baseline while ensuring montage consistency.
\end{abstract}

\section{Introduction}
Clinical electroencephalography (EEG) data are difficult to collect at scale. Recordings with clinically meaningful labels require expert review, and privacy constraints limit data sharing. Generative models offer a way to relieve this bottleneck, synthesizing data where labeled recordings are scarce or cannot be shared. Yet EEG generation remains relatively underexplored despite recent advances in diffusion and flow-based modeling~\citep{ddpm,lipman2023flow,Peebles_2023_ICCV}. The few existing approaches adapt generative adversarial networks~\citep{eeggan}, diffusion models~\citep{eegdiffusion} and flow matching models~\citep{jet}, typically treating EEG as a generic multichannel time series. This treatment overlooks a defining property of bipolar EEG: its channels are structurally constrained by how they are derived.

Clinical EEG is recorded against a common reference, but often read as potential differences between pairs of scalp electrodes. The chosen set of pairs, called a \textit{montage}, links electrodes into chains. Summing the differences along a chain cancels the intermediate electrodes and leaves only the difference between its endpoints. Two chains with the same endpoints must therefore sum to the same value at every timepoint. The montage used across our benchmarks contains sixteen electrode pairs and two independent loops. These impose two independent linear constraints, restricting the sixteen channel values at each timepoint to a fourteen-dimensional subspace. Real recordings satisfy these identities exactly, but samples synthesized by current generators do not.

Standard evaluation metrics do not explicitly measure montage violations. More importantly, improved distributional fidelity does not guarantee montage consistency. We therefore enforce these identities by construction rather than by penalizing violations. Specifically, we parameterize the model output within the subspace permitted by the montage. Invalid signals are thus excluded without introducing an additional tuning hyperparameter.

Enforcing the montage identities guarantees montage consistency, but does not by itself ensure fidelity to real EEG. Capturing dominant low-frequency activity while preserving low-energy transients and heavy-tailed amplitudes remains challenging. JET~\citep{jet} addresses waveform fidelity through explicitly specified auxiliary temporal, spectral, and statistical objectives. We instead use learned window-level conditioning to help preserve these characteristics. Specifically, we encode each multichannel EEG window into a single physiological state vector and condition the flow-matching decoder on it. This vector provides the decoder with window-specific context to guide generation. At sampling time, it is drawn from a prior fitted in the encoder’s latent space.

In summary, we propose \textbf{MontageFlow}, a flow-matching framework for montage-consistent, high-fidelity bipolar EEG generation that combines montage-constrained outputs with learned window-level conditioning. Across seven datasets, MontageFlow improves TS-FID by $1.3$--$5.7\times$ over the strongest baseline while satisfying the montage identities. Our contributions are summarized below:

\begin{itemize}
    \item \textbf{Montage-consistent bipolar EEG generation.} We show that state-of-the-art generators synthesize montage-inconsistent EEG signals and introduce the \textit{montage index} to measure this ($0$ = consistent). \ours{} restricts its output to the subspace of valid signals (Section~\ref{sec:montageflow}).
    \item \textbf{Conditioning mitigates oversmoothing and validity improves fidelity.} We observe that latent conditioning mitigates oversmoothing of EEG. We also find that enforcing validity improves fidelity, lowering TS-FID on all seven datasets by $10$ to $54$ points (Section~\ref{sec:ablations}).
    \item \textbf{State-of-the-art results.} Across seven EEG datasets, \ours{} improves TS-FID by $1.3$--$5.7\times$ over the strongest baseline, with a montage index of exactly $0$ (Section~\ref{sec:results}).
\end{itemize}

\section{Preliminaries}
\label{sec:prelim}
We first derive the exact linear constraints imposed by the bipolar EEG montage, then show that standard flow matching does not guarantee that generated outputs satisfy them.

\begin{figure}[t]
  \centering
  \includegraphics[width=\linewidth]{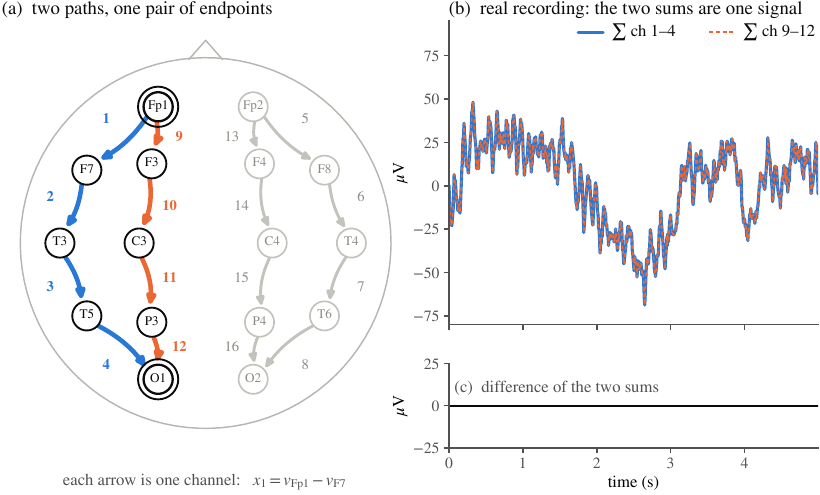}
  \caption{\textbf{The bipolar montage induces two exact linear
  constraints.}
  \textbf{(a)} Each channel is an oriented difference between two electrode
  potentials. Channels 1--4 and 9--12 form two paths from Fp1 to O1, and
  channels 5--8 and 13--16 do the same from Fp2 to O2. Either path
  sums to the potential difference between its endpoints.
  \textbf{(b)} The two left path sums overlap on a real TUSZ window, and
  \textbf{(c)} their difference is numerically zero.
  }
  \label{fig:montage}
\end{figure}

\noindent\textbf{Bipolar montage geometry.}
Let
\(
X=[\mathbf{x}(\tau_1),\ldots,\mathbf{x}(\tau_L)]
\in\mathbb{R}^{C\times L}
\)
denote an EEG window with \(C=16\) channels and \(L\) physical time points.
A bipolar montage represents channel \(i\) as the potential difference
between an ordered pair of electrodes:
\begin{equation}
  x_i(\tau)
  =
  v_{a_i}(\tau)-v_{b_i}(\tau),
  \qquad i=1,\ldots,C .
  \label{eq:bipolar}
\end{equation}
As illustrated in Figure~\ref{fig:montage}(a), summing channels along a
directed path cancels the potentials at intermediate electrodes, leaving
only the difference between the path endpoints. The montage contains two
pairs of paths with common endpoints. On the left hemisphere, channels
\(1\)--\(4\) and \(9\)--\(12\) both connect \(\mathrm{Fp1}\) to
\(\mathrm{O1}\). On the right hemisphere, channels \(5\)--\(8\) and
\(13\)--\(16\) both connect \(\mathrm{Fp2}\) to \(\mathrm{O2}\).
Consequently, the following equalities hold at every physical time point
\(\tau\):
\begin{align}
  x_1(\tau)+x_2(\tau)+x_3(\tau)+x_4(\tau)
  &=
  x_9(\tau)+x_{10}(\tau)+x_{11}(\tau)+x_{12}(\tau),
  \label{eq:left_constraint}\\
  x_5(\tau)+x_6(\tau)+x_7(\tau)+x_8(\tau)
  &=
  x_{13}(\tau)+x_{14}(\tau)+x_{15}(\tau)+x_{16}(\tau).
  \label{eq:right_constraint}
\end{align}
Figure~\ref{fig:montage}(b) verifies Eq.~\ref{eq:left_constraint} on a representative TUSZ window.

\noindent\textbf{Montage constraint.}
Let \(\mathbf{1}_4\) and \(\mathbf{0}_4\) denote four-dimensional vectors
of ones and zeros, respectively. The two path equalities can be encoded by
the constraint matrix
\begin{equation}
  A
  =
  \begin{bmatrix}
    \mathbf{1}_4^\top & \mathbf{0}_4^\top
      & -\mathbf{1}_4^\top & \mathbf{0}_4^\top \\
    \mathbf{0}_4^\top & \mathbf{1}_4^\top
      & \mathbf{0}_4^\top & -\mathbf{1}_4^\top
  \end{bmatrix}
  \in\mathbb{R}^{2\times16},
  \label{eq:constraint_matrix}
\end{equation}
where the four column groups correspond to channels \(1\)--\(4\),
\(5\)--\(8\), \(9\)--\(12\), and \(13\)--\(16\), respectively. Thus, \(A\mathbf{x}(\tau)=0\) at every physical time point. Since \(A\)
acts along the channel dimension, the corresponding window-level constraint
is
\begin{equation}
  AX=0_{2\times L}.
  \label{eq:constraint}
\end{equation}
The set of montage-valid channel vectors is therefore
\(
\ker A=\{\mathbf{u}\in\mathbb{R}^{16}:A\mathbf{u}=0\}.
\)
Since the two rows of \(A\) are linearly independent,
\(\operatorname{rank}(A)=2\), and \(\ker A\) is a 14-dimensional subspace
of \(\mathbb{R}^{16}\). 

\noindent\textbf{Standard flow matching.}
Let \(E\in\mathbb{R}^{C\times L}\) have i.i.d.\ standard Gaussian entries
and \(t\in[0,1]\) denote flow time. Linear flow
matching interpolates
\(X_t=tX+(1-t)E\) and trains a velocity network \(u_\theta(X_t,t)\) to
regress the target velocity \(X-E\). Generation starts from \(\widehat X_0=\varepsilon\sim\mathcal{N}(0,I)\) and integrates the learned
velocity along its own trajectory \(\widehat X_t\): \(  \widehat X =
  \varepsilon+\int_0^1 u_\theta(\widehat X_t,t)\,dt\). Applying the montage constraint of Eq.~\ref{eq:constraint} to this
sample gives
\begin{equation}
  A\widehat X
  =
  A\varepsilon+\int_0^1 A\,u_\theta(\widehat X_t,t)\,dt
  \label{eq:fm_violation}
\end{equation}
where \(A\varepsilon\neq0\) almost surely and \(u_\theta\) is unconstrained. Nothing
forces the two terms to cancel, so standard flow matching does not
guarantee montage-valid EEG.

\section{MontageFlow}
\label{sec:montageflow}

\begin{figure}[t]
  \centering
  \includegraphics[width=0.95\linewidth]{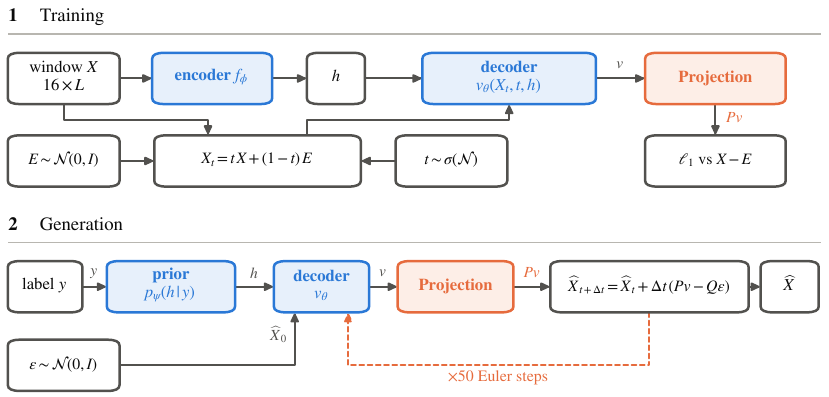}
  \caption{\textbf{Overview of MontageFlow.}
  During training, an encoder \(f_\phi\) maps an EEG window \(X\) to a
  global latent \(h\). The decoder receives the current flow state \(X_t\),
  flow time \(t\), and perturbed latent \(\widetilde h\), 
	and predicts a velocity field. After encoder-decoder training,
  a class-conditional rectified-flow prior \(p_\psi(h\mid y)\) is fitted to
  the encoded training latents and replaces the encoder during generation. 
	The predicted velocity is projected onto \(\ker A\). 
	At inference, the part of the source outside \(\ker A\) decays to zero, as in training.}
  \label{fig:method}
\end{figure}

\noindent\textbf{Overview.}
The overview of MontageFlow is shown in Figure~\ref{fig:method}. MontageFlow removes both sources of montage
violation (Eq.~\ref{eq:fm_violation}) by construction:
the velocity network's output is always projected onto \(\ker A\), and the starting noise outside \(\ker A\) decays to zero.
For fidelity, the velocity network is additionally conditioned on a global
latent encoding of the target window.

\noindent\textbf{Montage projector.}
Projecting the noise and the velocity onto \(\ker A\) requires the same
operation: given any channel vector \(\mathbf{u}\in\mathbb{R}^{16}\),
return a vector in \(\ker A\). The vector \(A\mathbf{u}\in\mathbb{R}^{2}\) holds the amounts by which
\(\mathbf{u}\) violates Eqs.~\ref{eq:left_constraint}
and~\ref{eq:right_constraint} respectively. To make \(\mathbf{u}\) valid, we cancel each violation by spreading it
equally over the eight channels involved. Each of the four channels on the
left-hand side of the equation is decreased by one eighth of the
violation, and each of the four on the right-hand side is increased by the
same amount: 

\begin{equation}
  P\mathbf{u}
  =
  \mathbf{u}-\tfrac{1}{8}A^\top(A\mathbf{u}),
  \qquad
  P
  =
  I-\tfrac{1}{8}A^\top A .
  \label{eq:projector}
\end{equation}
Since each row of \(A\) has eight entries equal to \(\pm1\) and the two
rows share no channel, \(AA^\top=8I_2\). Hence,
\(AP=A-\tfrac{1}{8}(AA^\top)A=0\). Applied to a window \(X\), \(P\) acts on each
column, so \(A(PX)=(AP)X=0\). Therefore, $P$ maps every window to a montage-valid window.

\noindent\textbf{Projected velocity.}
Let \(v_\theta(X_t,t,h)\in\mathbb{R}^{C\times L}\) be the raw output of
the velocity network, conditioned on flow time \(t\)
and a latent \(h\) introduced below. MontageFlow uses the projected
velocity
\begin{equation}
  u_\theta(X_t,t,h)
  =
  P\,v_\theta(X_t,t,h),
  \label{eq:projected_velocity}
\end{equation}
which satisfies \(A u_\theta=0\).
We sample \(t=\operatorname{sigmoid}(z)\), with
\(z\sim\mathcal{N}(-0.8,0.8^2)\), and set \(X_t=tX+(1-t)E\).
We optimize the following \(\ell_1\) velocity-matching objective:
\begin{equation}
  \mathcal{L}_{\mathrm{MF}}(\theta,\phi)
  =
  \mathbb{E}_{X,E,t,\eta,\sigma}
  \left[
    \left\|
      (X-E)
      -
      P\,v_\theta(X_t,t,\widetilde h)
    \right\|_1
  \right].
  \label{eq:objective}
\end{equation}
Here, \(\widetilde h=f_\phi(X)+\sigma\eta\) is the perturbed
conditioning latent, defined in Eq.~\ref{eq:latent_perturbation}.

\noindent\textbf{Validity by construction.}
The remaining source of violation is the start state \(\widehat X_0=\varepsilon\sim\mathcal{N}(0,I)\), unprojected as in training. With \(Q=I-P\), its part outside \(\ker A\), \((1-t)Q\varepsilon\), vanishes at \(t=1\):
\begin{equation}
  \widehat X
  =
  P\varepsilon+\int_0^1 P\,v_\theta(\widehat X_t,t,h)\,dt .
  \label{eq:montageflow_sample}
\end{equation}
As \(AP=0\), applying \(A\) gives \(A\widehat X=AP\varepsilon+\int_0^1 AP\,v_\theta(\widehat X_t,t,h)\,dt=0\), satisfying Eq.~\ref{eq:constraint}.

\noindent\textbf{Latent conditioning.}
Clinical EEG varies widely from window to window in amplitude and rhythm.
Generators trained with a regression loss can average this variety away
and produce oversmoothed samples. To mitigate this, we give the velocity
network a global view of the window it is generating by conditioning it on
a latent physiological state vector:
\begin{equation}
  h=f_\phi(X)\in\mathbb{R}^{256},
  \label{eq:latent}
\end{equation}
computed by an encoder \(f_\phi\) from the clean target window \(X\)
during training. At generation time, the conditioning vector is sampled from a learned
latent prior rather than obtained by encoding a real window.
To make the decoder robust to deviations between prior samples and
encoded latents, we perturb the latent during training:
\begin{equation}
  \widetilde h=f_\phi(X)+\sigma\eta,
  \qquad
  \eta\sim\mathcal{N}(0,I_{256}),
  \quad
  \sigma\sim\mathcal{U}(0,1).
  \label{eq:latent_perturbation}
\end{equation}
The velocity network receives \(\widetilde h\) as conditioning
in Eq.~\ref{eq:objective}, and the encoder is trained jointly with it.

\noindent\textbf{Latent prior.}
After jointly training the encoder and the velocity network, we replace
the encoder with a learned prior. We encode the training set into a latent bank
\(\mathcal{H}=\{(f_\phi(X_i),y_i)\}_{i=1}^{n}\) and fit a class-conditional
rectified-flow prior \(p_\psi(h\mid y)\) to these pairs.
The velocity network is not conditioned on \(y\), so the label enters
generation only through the latent.

\noindent\textbf{Sampling.}
To generate a window, we draw a class \(y\), a latent
\(h\sim p_\psi(h\mid y)\), and a start state \(\widehat X_0=\varepsilon\),
then integrate Eq.~\ref{eq:montageflow_sample} with \(K\) Euler steps.

\section{Experimental setup}
\label{sec:setup}
\noindent\textbf{Datasets.} We evaluate experiments on TUAB~\citep{tuheeg}, TUEV~\citep{tuheeg}, TUSZ~\citep{tusz}, CHB-MIT~\citep{shoeb2009chbmit}, NMT~\citep{khan2022nmt}, Siena~\citep{detti2020siena} and Helsinki~\citep{stevenson2019neonatal}. Details are in Appendix~\ref{app:datasets}.

\noindent\textbf{Metrics.} Our primary fidelity metric is \textbf{(a) TS-FID} (lower is better): the Fr\'echet distance between
representations of real and generated multivariate signals. It is
computed from FFT magnitudes, so it sees spectral structure but is blind to montage consistency. Hence, we propose the \textbf{(b) montage index}, which measures montage consistency: $0$ represents perfect consistency, $1$ is typical of independent channels. We also use \textbf{(c) shape correlation}, which compares the
generated cross-channel correlation structure against the real one (higher is better). Finally, we report \textbf{(d) physiological metrics}: PSD slope, envelope, Hjorth activity, Hjorth mobility and drift (lower is better). All are defined in Appendix~\ref{app:metrics}.

\noindent\textbf{Baselines.} EEG-GAN~\citep{eeggan}, vanilla diffusion~\citep{ddpm}, vanilla flow matching~\citep{lipman2023flow}, JET~\citep{jet}, and its cross-channel variant JET$^{\times}$.

\noindent\textbf{Implementation.} The encoder of Section~\ref{sec:montageflow} is a causal Transformer (8 blocks, width 512, 8 heads). The encoder output is projected to a single 256-dimensional latent $h$. The decoder is a 1-D DiT-B transformer (12 blocks, width 768, 12 heads) with adaLN conditioning on the timestep and $h$. Each channel is split into patches of size $100$, and attention runs over the patches of all channels. To stabilize training, we apply QK normalization in the decoder's attention. Together they account for $156$M
parameters, of which the decoder holds $130$M. Training runs for $100$k steps at a batch of $128$. We use
AdamW with a learning rate of $5\times10^{-5}$. The prior is then fit on the encoded training split for $\sim10$k steps, and sampling (Figure~\ref{fig:method}) uses $50$ Euler steps. We evaluate $10{,}000$ generated windows against $10{,}000$ windows of the training split.

\section{Results}
\label{sec:results}
\begin{table}[!h]
  \centering\footnotesize\setlength{\tabcolsep}{2.6pt}
  \caption{Generation quality and montage consistency (\textbf{bold}: best, \underline{underline}: second best).}
  \label{tab:main}
  \begin{tabular}{lcccccccc}
\toprule
\multirow{2}{*}{Method} & \multicolumn{7}{c}{TS-FID $\downarrow$} & \multicolumn{1}{c}{Montage index $\downarrow$} \\
\cmidrule(lr){2-8}\cmidrule(lr){9-9}
 & TUSZ & TUEV & TUAB & CHB-MIT & NMT & Siena & Helsinki & Avg. \\
\midrule
EEG-GAN & $358.6$ & $315.1$ & $394.3$ & $275.2$ & $194.5$ & $428.5$ & $277.6$ & $0.486$ \\
Vanilla diffusion & $\underline{135.1}$ & $165.7$ & $\underline{147.0}$ & $\underline{88.3}$ & $\underline{51.4}$ & $\underline{187.7}$ & $\underline{147.6}$ & $\underline{0.131}$ \\
Vanilla flow matching & $147.3$ & $\underline{142.2}$ & $181.8$ & $132.1$ & $89.5$ & $198.4$ & $190.4$ & $0.351$ \\
JET & $218.4$ & $229.6$ & $257.1$ & $178.8$ & $199.1$ & $253.9$ & $200.7$ & $0.974$ \\
JET$^{\times}$ & $221.1$ & $186.8$ & $234.5$ & $188.7$ & $193.5$ & $256.1$ & $189.7$ & $0.160$ \\
\midrule
\rowcolor{oursblue!9}\textbf{\ours{}} & $\mathbf{23.9}$ & $\mathbf{56.7}$ & $\mathbf{42.4}$ & $\mathbf{16.6}$ & $\mathbf{39.6}$ & $\mathbf{43.9}$ & $\mathbf{58.8}$ & $\mathbf{0.000}$ \\
\bottomrule
\end{tabular}

\end{table}

\noindent\textbf{Generation quality and montage consistency.}
\ours{} achieves the lowest TS-FID on all seven datasets and scores $1.3$--$5.7\times$ lower than the strongest baseline on each (Table~\ref{tab:main}). In every setting, the strongest baseline is a vanilla diffusion or flow-matching model sharing our backbone. Vanilla flow matching differs from \ours{} only in latent conditioning and montage projection, providing a controlled comparison of their combined contribution. Their individual effects are examined in Table~\ref{tab:ladder}.  \ours{} achieves a montage index of exactly $0$ on every dataset. JET averages $0.97$, while adding cross-channel attention (JET$^{\times}$) reduces it to $0.16$. Vanilla diffusion achieves the lowest index among the baselines, at $0.13$, but still violates the montage identities.

\noindent\textbf{Channel relationship.}
Montage validity alone does not guarantee realistic cross-channel
relationships: any projected output satisfies the montage identities,
even if its remaining channel dependencies differ from those of real
EEG. We therefore evaluate shape correlation
on TUSZ, TUEV and TUAB. \ours{} is closest to the real-data reference
on all three corpora, achieving $0.989$--$0.994$. The vanilla models, which also use
cross-channel attention, achieve $0.95$--$0.97$ but consistently
remain below \ours{}. JET scores near
$0$ across all settings: it generates each channel on its own, so its
samples miss the cross-channel structure of real EEG. Adding cross-channel
attention in JET$^{\times}$ raises the scores to $0.93$--$0.95$,
which suggests that channel dependencies are worth modeling explicitly.
EEG-GAN achieves substantially lower scores of $0.43$--$0.65$.

\begin{table}[t]
  \centering\footnotesize\setlength{\tabcolsep}{5.88pt}
  \caption{Downstream balanced accuracy ($\uparrow$, \textbf{bold}: best, \underline{underline}: second best).}
  \label{tab:downstream}
  \begin{tabular}{lcccccccc}
\toprule
Method & TUSZ & TUEV & TUAB & CHB-MIT & NMT & Siena & Helsinki & Avg. \\
\midrule
EEG-GAN & $0.485$ & $0.231$ & $0.502$ & $0.756$ & $0.508$ & $0.512$ & $0.495$ & $0.498$ \\
Vanilla diffusion & $0.614$ & $\underline{0.458}$ & $0.577$ & $\mathbf{0.877}$ & $\underline{0.666}$ & $\mathbf{0.662}$ & $\mathbf{0.588}$ & $\underline{0.634}$ \\
Vanilla flow matching & $0.613$ & $0.427$ & $0.551$ & $\underline{0.786}$ & $0.568$ & $0.513$ & $0.528$ & $0.569$ \\
JET & $\underline{0.744}$ & $0.421$ & $0.583$ & $0.628$ & $0.621$ & $0.590$ & $\underline{0.557}$ & $0.592$ \\
JET$^{\times}$ & $0.568$ & $0.448$ & $\underline{0.605}$ & $0.625$ & $0.570$ & $0.504$ & $0.513$ & $0.548$ \\
\midrule
\rowcolor{oursblue!9}\textbf{\ours{}} & $\mathbf{0.779}$ & $\mathbf{0.624}$ & $\mathbf{0.742}$ & $0.678$ & $\mathbf{0.697}$ & $\underline{0.611}$ & $0.504$ & $\mathbf{0.662}$ \\
\bottomrule
\end{tabular}

\end{table}

\noindent\textbf{Downstream utility.}
We train the same classifier (CBraMod;~\citealp{cbramod}) for every method on $5{,}000$ generated windows per class and test it on real windows (Table~\ref{tab:downstream}).
We select the best epoch on the real validation split and average over three seeds.
As a reference, the same classifier trained on $5{,}000$ real windows per class reaches $0.754$ on average.
\ours{} achieves the highest average and is best on four of the seven datasets.

\section{Ablations and Analyses}
\label{sec:ablations}

\begin{wraptable}{r}{0.5\linewidth}
  \vspace{-12pt}
  \centering\footnotesize\setlength{\tabcolsep}{3pt}
  \caption{Incremental ablation from vanilla flow matching to \ours{} (TS-FID~$\downarrow$).}
  \label{tab:ladder}
  \begin{tabular}{lc}
\toprule
Method & TUAB \\
\midrule
Vanilla flow matching ($x$-parameterization) & $247.8$ \\
$+$ velocity parameterization & $194.9$ \\
$+$ cross-channel attention & $181.8$ \\
$+$ latent conditioning & $97.3$ \\
$+$ latent perturbation & $96.6$ \\
\midrule
\rowcolor{oursblue!9}$+$ montage projection & $\mathbf{42.4}$ \\
\bottomrule
\end{tabular}

  \vspace{-10pt}
\end{wraptable}

\paragraph{Incremental ablation with respect to vanilla flow matching.}
We present an incremental ablation on TUAB in Table~\ref{tab:ladder}, which starts from flow matching and adds one component per row until it reaches \ours{}.
Switching from $x$- to velocity parameterization lowers TS-FID by $53$.
Adding cross-channel attention gives the vanilla flow matching baseline of Table~\ref{tab:main}.
It lowers TS-FID by $13$.
Latent conditioning (row~4) lowers TS-FID by a further $85$, the largest gain in the ladder. Perturbing the latent during training (row~5) lowers it by less than $1$.
Finally, the projection makes every generated EEG window montage-consistent.
It also improves distributional fidelity, lowering TS-FID by a further $54$ points.

\noindent\textbf{Is post-hoc projection enough?}
The projection could instead be applied only at inference, to the model trained without it (row~5 of Table~\ref{tab:ladder}).
Both approaches make every generated window montage-consistent, but they differ sharply in fidelity.
Compared with post-hoc projection, training with it lowers TS-FID by $13$--$86$ points and the average physiological distance on all seven datasets (Table~\ref{tab:postproj}).
PCFM~\citep{utkarsh2025pcfm} instead enforces the constraint while sampling the same model.
Its samples are also montage-consistent, but \ours{} beats it on TS-FID on six of the seven datasets and on the physiological metrics on all seven.
We hypothesize that the unconstrained model wastes modeling capacity on invalid signals, while \ours{} does not.

\begin{figure}[t]
  \centering
  \includegraphics[width=\linewidth]{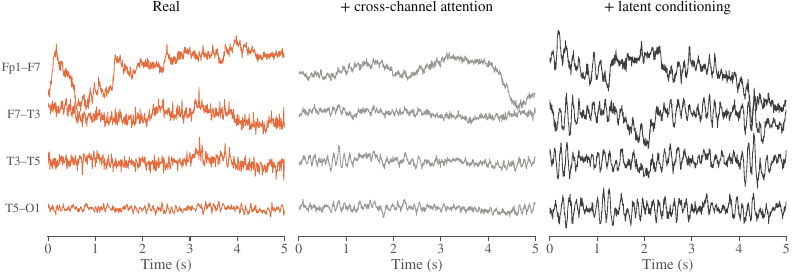}
  \caption{\textbf{Latent conditioning mitigates oversmoothing.} Representative samples from TUAB are shown.}
  \label{fig:paired}
\end{figure}

\begin{table}[!h]
  \centering\small\setlength{\tabcolsep}{4pt}
  \caption{\textbf{Is post-hoc projection enough?} PCFM, no projection and inference only share one model trained without projection. Physio.: mean of the five physiological metrics (\textbf{bold}: best, \underline{underline}: second best).}
  \label{tab:postproj}
  \resizebox{\linewidth}{!}{\begin{tabular}{lcccccc>{\columncolor{oursblue!9}}c>{\columncolor{oursblue!9}}c}
\toprule
 & \multicolumn{2}{c}{PCFM} & \multicolumn{2}{c}{No projection} & \multicolumn{2}{c}{Inference only} & \multicolumn{2}{c}{Training $+$ inference} \\
\cmidrule(lr){2-3}\cmidrule(lr){4-5}\cmidrule(lr){6-7}\cmidrule(lr){8-9}
Dataset & TS-FID $\downarrow$ & Physio. $\downarrow$ & TS-FID $\downarrow$ & Physio. $\downarrow$ & TS-FID $\downarrow$ & Physio. $\downarrow$ & TS-FID $\downarrow$ & Physio. $\downarrow$ \\
\midrule
TUSZ & \underline{37.5} & \underline{0.271} & 46.3 & 0.325 & 50.6 & 0.308 & \textbf{23.9} & \textbf{0.258} \\
TUEV & 84.9 & 0.346 & 80.9 & 0.320 & \underline{80.1} & \underline{0.297} & \textbf{56.7} & \textbf{0.231} \\
TUAB & 112.6 & 0.540 & \underline{96.6} & \underline{0.226} & 128.2 & 0.258 & \textbf{42.4} & \textbf{0.146} \\
CHB-MIT & 55.3 & 0.464 & \underline{26.7} & \underline{0.105} & 29.9 & 0.117 & \textbf{16.6} & \textbf{0.087} \\
NMT & \textbf{23.6} & 0.921 & 70.7 & 0.395 & 88.2 & \underline{0.327} & \underline{39.6} & \textbf{0.312} \\
Siena & 81.0 & 0.383 & \underline{78.2} & \underline{0.185} & 95.2 & 0.197 & \textbf{43.9} & \textbf{0.144} \\
Helsinki & \underline{101.3} & 0.562 & 103.4 & 0.410 & 112.5 & \underline{0.361} & \textbf{58.8} & \textbf{0.266} \\
\midrule
Avg. & \underline{70.9} & 0.498 & 71.8 & 0.281 & 83.5 & \underline{0.267} & \textbf{40.3} & \textbf{0.206} \\
\bottomrule
\end{tabular}
}
\end{table}
\noindent\textbf{Latent conditioning mitigates oversmoothing.}
Figure~\ref{fig:paired} shows four channels of real EEG (left column) and of generated EEG (middle and right columns). The two generated columns differ only in latent conditioning, isolating its effect on oversmoothing. Without the latent, generated windows collapse to a flat average and keep $42$--$50\%$ of the real median amplitude. With it, they keep $67$--$84\%$.

\noindent\textbf{The regression loss ($L_1$ versus $L_2$).}
Flow matching conventionally uses squared-error velocity regression.
Compared with $L_2$, $L_1$ places less emphasis on large residuals.
We evaluate this choice by training \ours{} under each loss on TUSZ.
Using $L_2$ instead of $L_1$ raises TS-FID from $23.9$ to $41.8$.
We therefore adopt $L_1$ as the training objective.

\noindent\textbf{Deriving bipolar EEG from common-reference EEG.}
Montage consistency also holds if a model generates the common-reference electrode signals and the bipolar channels are derived from them afterward.
We train \ours{} without projection on the common-reference recordings of TUH (TUSZ, TUEV, TUAB) and score the derived bipolar windows.
This route is montage-consistent too, but less faithful: compared with \ours{}, TS-FID rises from $23.9$ to $57.1$ on TUSZ, from $56.7$ to $57.9$ on TUEV and from $42.4$ to $80.9$ on TUAB.
The average physiological distance also rises substantially on all three datasets.
Generating the bipolar channels directly, through the projection, is therefore the better choice.

\begin{wraptable}{r}{0.47\linewidth}
  \vspace{-12pt}
  \centering\footnotesize\setlength{\tabcolsep}{2pt}
  \caption{Common-average EEG on TUH. Physio.: mean of the five physiological metrics.}
  \label{tab:car}
  \begin{tabular}{lcc>{\columncolor{oursblue!9}}c>{\columncolor{oursblue!9}}c}
\toprule
 & \multicolumn{2}{c}{No projection} & \multicolumn{2}{c}{Projection} \\
\cmidrule(lr){2-3}\cmidrule(lr){4-5}
Dataset & TS-FID $\downarrow$ & Physio. $\downarrow$ & TS-FID $\downarrow$ & Physio. $\downarrow$ \\
\midrule
TUSZ & 42.1 & 0.279 & \textbf{27.4} & \textbf{0.257} \\
TUEV & 55.9 & 0.277 & \textbf{45.3} & \textbf{0.202} \\
TUAB & 88.7 & 0.216 & \textbf{58.4} & \textbf{0.199} \\
\bottomrule
\end{tabular}

  \vspace{-10pt}
\end{wraptable}

\noindent\textbf{Applicability to other montages.}
The projection is not limited to bipolar montages.
In common-average EEG, the channels sum to zero at every time step, which is also a linear constraint.
We train \ours{} on the common-average recordings of TUH, with and without projection onto this constraint (Table~\ref{tab:car}, Appendix~\ref{app:car}).
The projection lowers TS-FID from $42.1$ to $27.4$ on TUSZ, from $55.9$ to $45.3$ on TUEV and from $88.7$ to $58.4$ on TUAB.
It also lowers the average physiological distance on all three datasets.

\begin{wraptable}{r}{0.4\linewidth}
  \vspace{-12pt}
  \centering\footnotesize\setlength{\tabcolsep}{2.5pt}
  \caption{Computational cost of training and generation across methods on TUAB.}
  \label{tab:cost}
  \begin{tabular}{lcc}
\toprule
Method & s/step & ms/window \\
\midrule
EEG-GAN & 2.39 & 2 \\
Vanilla diffusion & 0.17 & 74 \\
Vanilla flow matching & 0.21 & 78 \\
JET & 0.14 & 40 \\
JET$^{\times}$ & 0.11 & 37 \\
\midrule
\rowcolor{oursblue!9}\textbf{\ours{}} & 0.23 & 81 \\
\bottomrule
\end{tabular}

  \vspace{-10pt}
\end{wraptable}

\noindent\textbf{Computational cost.}
We report the computational cost of every method on TUAB in Table~\ref{tab:cost}. Vanilla diffusion and vanilla flow matching share the backbone, and vanilla flow matching is slightly slower per step ($0.21$ vs.\ $0.17$\,s). \ours{} adds an encoder with $26$M parameters, which raises the time per training step by about $10\%$. At generation time, sampling the latent from the prior adds about $2$\,ms per window, so it generates at nearly the speed of the vanilla models ($81$ vs.\ $78$\,ms). JET and JET$^{\times}$ are cheaper because they use a coarser patch, and EEG-GAN generates fastest but trains slowest.

\begin{figure}[t]
  \centering
  \includegraphics[width=1.0\linewidth]{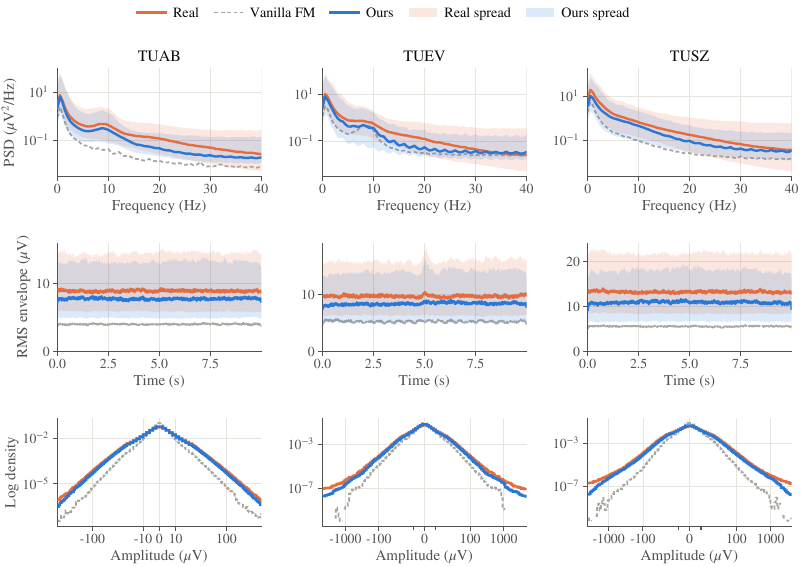}
  \caption{\textbf{Physiological diagnostics.} \textbf{Top:} power spectrum, median over windows, with shading from the 10th to the 90th percentile. \textbf{Middle:} amplitude over time (root-mean-square across channels), median over windows, with shading over the middle half of windows. \textbf{Bottom:} distribution of sample amplitudes (log density). Windows are 10\,s.}
  \label{fig:physio}
\end{figure}

\begin{table}[!t]
  \centering\small
  \caption{Physiological diagnostics, averaged over the seven datasets (\textbf{bold}: best, \underline{underline}: second best).}
  \label{tab:diag}
  \setlength{\tabcolsep}{10.82pt}\resizebox{\linewidth}{!}{\begin{tabular}{lccccc>{\columncolor{gray!12}}c}
\toprule
Method & PSD slope $\downarrow$ & Envelope $\downarrow$ & Hj.\ act. $\downarrow$ & Hj.\ mob. $\downarrow$ & Drift $\downarrow$ & Avg. $\downarrow$ \\
\midrule
EEG-GAN & 1.855 & 0.903 & 1.077 & 0.979 & 0.707 & 1.104 \\
Vanilla diffusion & 0.698 & 1.279 & 1.346 & 0.737 & \underline{0.149} & 0.842 \\
Vanilla flow matching & \underline{0.574} & 1.194 & 1.209 & 1.042 & 0.189 & 0.842 \\
JET & 0.818 & 0.580 & 0.606 & 0.759 & 0.437 & 0.640 \\
JET$^{\times}$ & 0.751 & \underline{0.463} & \underline{0.483} & \underline{0.620} & 0.210 & \underline{0.506} \\
\midrule
\rowcolor{oursblue!9}\textbf{\ours{}} & \textbf{0.269} & \textbf{0.183} & \textbf{0.204} & \textbf{0.319} & \textbf{0.057} & \textbf{0.206} \\
\bottomrule
\end{tabular}
}
\end{table}

\noindent\textbf{Physiological diagnostics.}
We report five more diagnostic metrics, averaged over the seven datasets, in Table~\ref{tab:diag} (Appendix~\ref{app:metrics}). \ours{} is closest to real EEG on every metric and on average, at $0.21$ against $0.51$ for the next best, JET$^{\times}$.
Furthermore, we qualitatively examine whether the generated signals behave like EEG by studying Figure~\ref{fig:physio}.
\textbf{(1)~Power spectrum:} The top row shows the power at each frequency. Real EEG has most of its power at the lowest frequencies and falls off with frequency, with an alpha peak near $8$--$9$\,Hz on TUAB and TUEV. Below about $12$\,Hz, our spectrum follows the real shape and reproduces the alpha peak at the same frequency. Vanilla flow matching has less power at almost every frequency, and on TUEV both generators exaggerate the alpha peak. Above $12$\,Hz, our spectrum falls below the real one, most visibly on TUAB. However, real windows have much higher variance at these frequencies, and our median stays within the real shading throughout. \textbf{(2)~Amplitude over time:} The middle row tracks signal amplitude across the window. Real EEG keeps a flat median envelope with no drift, and so does \ours{}. Our envelope sits slightly below the real one, at $82$--$87\%$ of its level, and its spread across windows is similar. Vanilla flow matching has about half the real amplitude. On TUEV, the real envelope rises near $5$\,s. \ours{} reproduces this rise, while vanilla flow matching barely shows it. \textbf{(3)~\mbox{Heavy-tailed} distributions:} The amplitude distribution of clinical EEG is heavy-tailed. A generator that collapses toward average behavior loses the extremes, where transient events such as spikes live (bottom row). Our density follows the real curve from the sharp central peak down through the tails. Vanilla flow matching is too narrow, and its tails vanish by $1000\,\mu$V. Above $100\,\mu$V, our samples keep $57$--$64\%$ of the real probability mass, against $3$--$9\%$ for vanilla flow matching.

\section{Related works}

\noindent\textbf{EEG models.}
EEG is routinely used to monitor brain activity and is essential for epilepsy diagnosis~\citep{tatum2018clinical}. Deep EEG models are predominantly discriminative and task-specific. For instance, EEGNet uses a compact CNN for EEG classification~\citep{eegnet}, while EEG Conformer combines convolution with self-attention to capture long-range temporal dependencies~\citep{eegconformer}. \citet{gemein2020} compare deep and feature-based decoders for distinguishing pathological from normal recordings on TUAB, while \citet{tang2022gnn} and REST~\citep{rest2024} use graph neural networks for seizure detection and classification on TUSZ. Even pretrained EEG foundation models~\citep{bendr,biot,labram,cbramod,reve} typically require fine-tuning or linear probing on labeled data for downstream tasks~\citep{labram,cbramod,reve}. However, achieving expert-level detection requires large expert-annotated datasets~\citep{sparcnet}. Annotation typically relies on certified neurologists, whose expertise is in short supply~\citep{tusz}. Moreover, clinically significant target events are rare: seizures account for less than 0.1\% of the TUH EEG archive~\citep{tusz}. Data access is also constrained by patient privacy: even TUH, the largest public EEG corpus, required HIPAA de-identification before release~\citep{tuheeg}. Hand-crafted augmentation can expand training data, but its effectiveness is task-dependent, and some transformations may alter the underlying semantics~\citep{rommel2022augment}. Per-channel augmentation of bipolar channels may also violate montage identities (Eq.~\ref{eq:constraint}). Alternatively, synthetic seizure EEG can substitute for real seizures in detector training, yielding comparable performance with lower re-identification risk~\citep{pascual2021epilepsygan}. Despite extensive research on EEG-based predictive modeling, EEG generation remains comparatively underexplored.

\looseness=-1\noindent\textbf{Generative modeling.}
Generative modeling has advanced rapidly, driven largely by vision. GANs enabled high-fidelity, high-resolution image synthesis~\citep{goodfellow2014gan,stylegan}, but adversarial training can be unstable and prone to mode collapse~\citep{wgan,wgangp}. Diffusion models~\citep{sohl2015diffusion,ddpm,song2021sde} use a stable, non-adversarial training objective and have surpassed GANs in image quality~\citep{dhariwal2021beat}. Closely related to diffusion, flow matching offers an alternative approach to generative modeling by learning a velocity field from noise to data~\citep{lipman2023flow,liu2023rectified}. With Gaussian noise, the two differ mainly in how they weight noise levels~\citep{kingma2023understanding}. Architectures for both diffusion and flow-matching models have evolved from U-Nets~\citep{unet,ddpm} to transformers operating on latent patches~\citep{Peebles_2023_ICCV} and, more recently, on raw pixels while predicting the clean signal~\citep{jit}. For time series, GAN- and diffusion-based generators~\citep{timegan,diffusionts,vetter2024ddpm} use generic inductive biases without encoding domain-specific constraints. Constrained generative models handle general or inequality-defined feasible sets by intervening during sampling~\citep{christopher2024projected,utkarsh2025pcfm} or by modifying the diffusion process~\citep{lou2023reflected,liu2023mirror}. The montage constraint, by contrast, is a linear equality with a fixed, closed-form projector applied during both training and sampling. Among EEG generators, EEG-GAN~\citep{eeggan} uses progressive training with a modified WGAN-GP objective. Beyond GANs, \citet{eegdiffusion} use a diffusion model to synthesize preictal EEG for seizure prediction. Furthermore, JET~\citep{jet} trains a flow-matching transformer with hand-engineered auxiliary temporal and statistical losses. However, EEG-GAN synthesizes low-fidelity samples, as seen in Table~\ref{tab:main}, and is slow to train (Table~\ref{tab:cost}). JET needs its auxiliary-loss weights tuned for each dataset. None of them generates montage-consistent EEG. In contrast, \ours{} synthesizes high-fidelity EEG that is also montage-consistent.

\section{Conclusion}
We show that current bipolar EEG generators synthesize montage-inconsistent signals. MontageFlow rules out such signals by restricting its output to the subspace of valid signals. It further conditions the flow-matching decoder on a physiological state vector. With latent conditioning, enforcing validity during training also improves fidelity. Across seven publicly available EEG corpora, \ours{} improves TS-FID by $1.3$--$5.7\times$ over the strongest baseline with perfect montage consistency. More broadly, our results suggest that validity and fidelity need not trade off when known linear identities are built into the generator.

\subsection*{Limitations and future work}
Our study focuses on the bipolar montage, with one common-average montage experiment on TUH. Within this scope, we have not performed experiments on high-density or BCI recordings. In future work, \ours{} could be extended to other modalities with linear identities, such as ECG.

\subsection*{AI use statement}
In this work, we used generative AI tools for coding, drawing the diagrams, polishing the writing, finding related works and writing the \LaTeX{} code. We take responsibility for the final content of this work, including text, claims or artifacts produced with the aid of generative AI.

\ificlrfinal\subsection*{Acknowledgments}
The authors acknowledge support for this research from \textbf{Evergreen: A Generative AI \& Behavioral Sensing Digital Ecosystem to Promote Student Wellness and Flourishing.} This work is made possible through philanthropic gifts to Dartmouth College dedicated to advancing AI-supported well-being and flourishing of college students.
\fi %

\bibliography{iclr2027_conference}
\bibliographystyle{iclr2027_conference}

\newpage
\appendix
{\centering\Large\textbf{Appendix}\par}
\vspace{-0.5em}
\section{Datasets}
\label{app:datasets}

Table~\ref{tab:data} lists the corpora used in our study. TUAB (abnormal/normal) and TUSZ (seizure/background) have 2 classes each. TUEV has 6 event types and is heavily imbalanced. CHB-MIT, Siena and Helsinki are labeled seizure/background, and NMT abnormal/normal.

\begin{table}[H]
  \centering\small
  \caption{Corpora used in our study. Windows lie on the 16-channel bipolar montage at 200\,Hz.}
  \label{tab:data}
  \begin{tabular}{lcccc}
    \toprule
    & classes & window & train windows & test windows \\
    \midrule
    TUSZ & 2 & $16\times2000$ & 13{,}667 & 5{,}008 \\
    TUEV & 6 & $16\times2000$ & 68{,}445 & 29{,}421 \\
    TUAB & 2 & $16\times2000$ & 297{,}103 & 36{,}945 \\
    \midrule
    CHB-MIT & 2 & $16\times2000$ & 56{,}522 & 4{,}059 \\
    NMT & 2 & $16\times2000$ & 121{,}857 & 13{,}118 \\
    Siena & 2 & $16\times2000$ & 39{,}610 & 10{,}965 \\
    Helsinki & 2 & $16\times2000$ & 32{,}591 & 7{,}660 \\
    \bottomrule
  \end{tabular}
\end{table}

\paragraph{Common-average EEG.}\label{app:car}
Here the projection discards the channel mean. To keep training numerically stable, we flip the sign of half of the channels around the velocity network, $u_\theta=P\,S\,v_\theta(SX_t,t,\widetilde h)$ with $S=\operatorname{diag}(s)$, $s\in\{\pm1\}^{16}$ and $\mathbf{1}^\top s=0$.

\section{Metric definitions}
\label{app:metrics}

TS-FID, shape correlation and the physiological metrics compare $10{,}000$ generated windows with $10{,}000$ windows of the training split. The montage index uses only the generated windows.

\paragraph{TS-FID.}
The Fr\'echet distance between Gaussians fitted to the features $\phi(X)$ of real and generated windows,
\begin{equation*}
  \mathrm{TS\text{-}FID}=\lVert\mu_r-\mu_g\rVert_2^2+\operatorname{Tr}\!\big(\Sigma_r+\Sigma_g-2(\Sigma_r\Sigma_g)^{1/2}\big),
\end{equation*}
with $(\mu,\Sigma)$ the mean and covariance of the features over the $10{,}000$ windows. Following \citet{jet}, $\phi$ takes the magnitude of the discrete Fourier transform along time and keeps the lower half of the frequency bins. It average-pools frequency to at most $256$ bins and channels to $4$ groups, applies $\log(1+\cdot)$, and standardizes within the window.

\paragraph{Montage index.}
Each constraint in Eqs.~\ref{eq:left_constraint}--\ref{eq:right_constraint} says that two chains of channels between the same pair of electrodes must sum to the same signal. The \emph{residual} is the difference between the two sums. Within each window, we average its absolute value over time and divide by the root-mean-square of the $m=8$ channels involved. We average this over windows and both constraints, then divide by $\sqrt{2m/\pi}\approx2.257$, its value for independent Gaussian channels. So $0$ is montage-consistent and $1$ is typical of independent channels.

\paragraph{Shape correlation.}
We average the $16\times16$ per-window Pearson correlation matrix of the channels over windows. Shape correlation is the Pearson correlation between the off-diagonal entries of this matrix for generated and for real windows. It asks whether the same channel pairs are coupled.

\paragraph{Physiological metrics.}
Each metric in Table~\ref{tab:diag} is a per-window scalar $g(X)$. Its distributions over generated and real windows are compared by the $1$-Wasserstein distance, divided by the interquartile range of the real values (lower is better),
\begin{equation*}
  d(g)=W_1\big(P^{\mathrm{gen}}_g,\,P^{\mathrm{real}}_g\big)\,/\,\mathrm{IQR}\big(P^{\mathrm{real}}_g\big).
\end{equation*}

\paragraph{PSD slope.} The exponent $\chi$ of a least-squares fit $\log_{10}S(f)\approx a-\chi\log_{10}f$ to each channel's Welch spectrum over $1$--$40$\,Hz, averaged over channels. An oversmoothed signal has a steeper slope.

\paragraph{Envelope.} The $\log_{10}$ RMS over time of $e(t)=\big(\frac{1}{C}\sum_c x_c(t)^2\big)^{1/2}$. It measures how much energy a window carries.

\paragraph{Hjorth activity.} $\frac{1}{C}\sum_c\log_{10}\mathrm{var}(x_c)$, the per-channel counterpart of the envelope.

\paragraph{Hjorth mobility.} $\frac{1}{C}\sum_c\sqrt{\mathrm{var}(\dot x_c)/\mathrm{var}(x_c)}$, with $\dot x_c$ the first difference. It is scale-free, so it reflects frequency content rather than amplitude.

\paragraph{Drift.} The least-squares slope of $e(t)$ across the window, multiplied by $L$ and divided by the mean of $e(t)$. It detects windows whose amplitude ramps up or decays.

\section{Qualitative samples}
\label{app:qualitative}
Real EEG mixes an ongoing background rhythm with brief sharp events. \ours{} reproduces the background rhythm on all three corpora, though its traces are quieter than real on TUEV and TUSZ. The baselines fail differently. Vanilla diffusion and vanilla flow matching are far too quiet. JET's traces are over-regular: they rise and fall with a fixed rhythm.

\begin{figure}[H]
  \centering
  \includegraphics[width=0.85\linewidth]{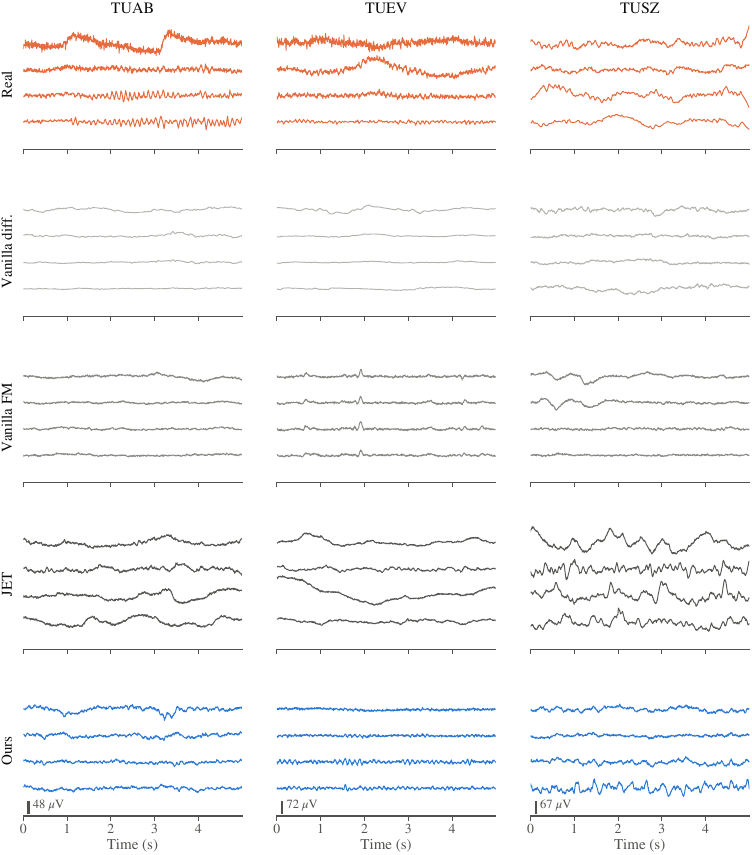}
  \caption{\textbf{Generated and real windows.} First four channels of the bipolar montage, five seconds, one window per corpus and method (the median-amplitude window among $2{,}000$ samples).}
  \label{fig:qualitative}
\end{figure}
\clearpage

\section{Channel-wise analysis}
\label{app:channelwise}

\begin{figure}[H]
  \centering
  \includegraphics[width=\linewidth,height=0.72\textheight,keepaspectratio]
    {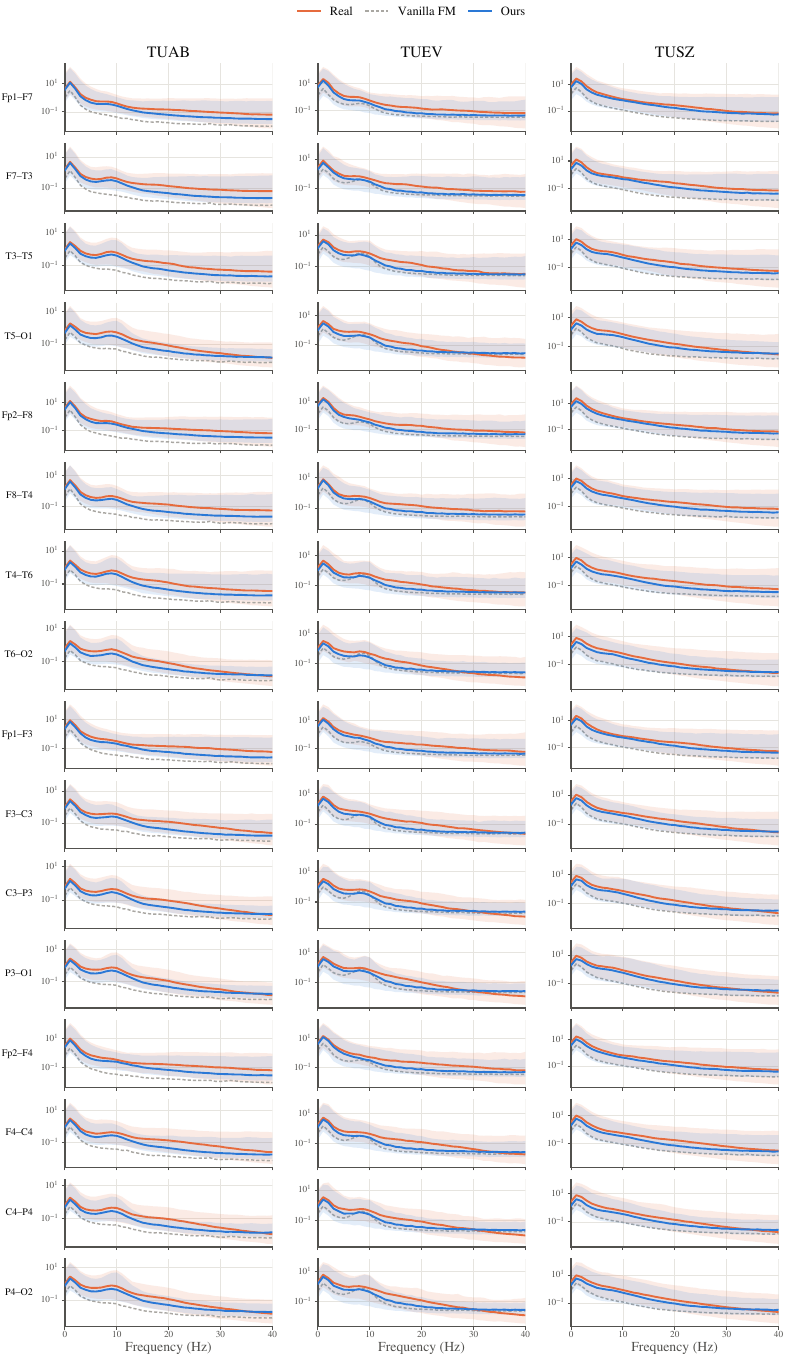}
  \caption{\textbf{Channel-wise power spectrum.} Median over windows for each channel, with shading from the 10th to the 90th percentile.}
  \label{fig:chan-psd}
\end{figure}

\begin{figure}[H]
  \centering
  \includegraphics[width=\linewidth,height=0.85\textheight,keepaspectratio]
    {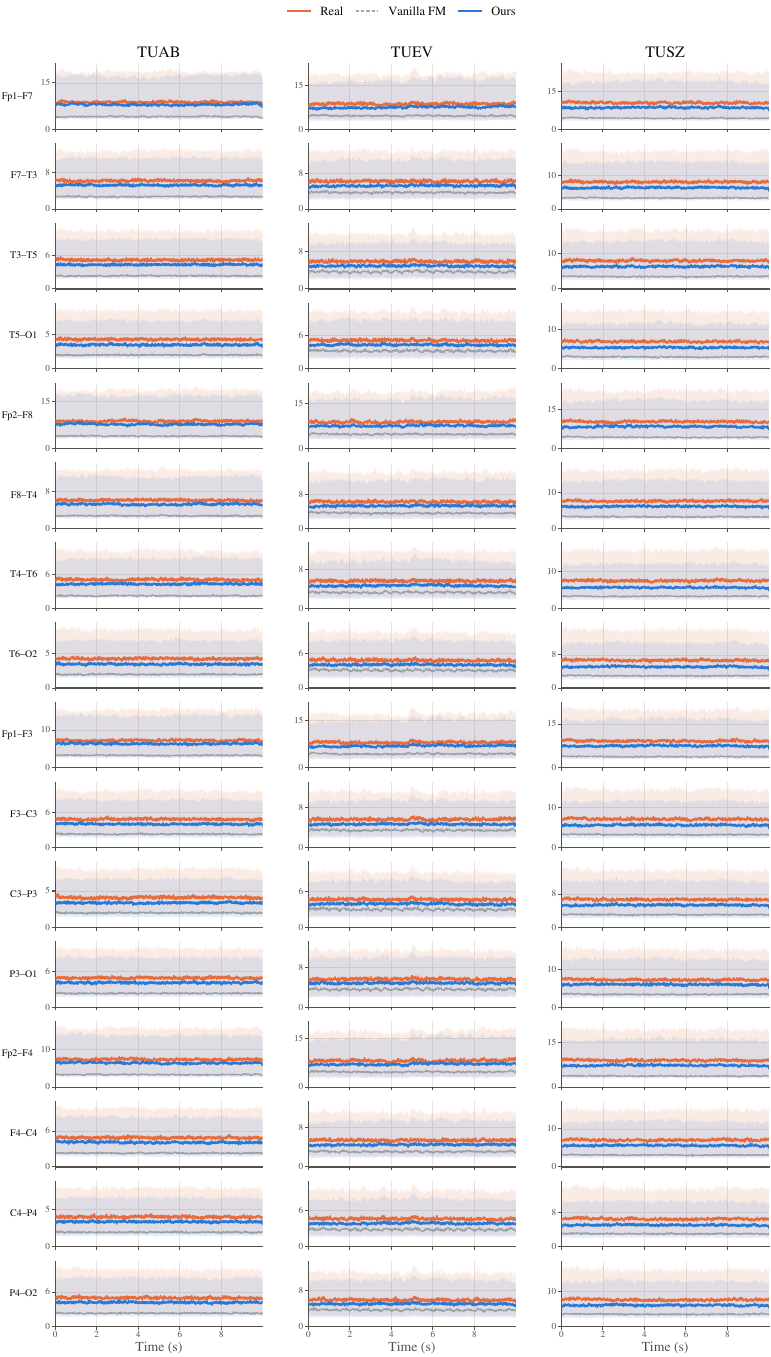}
  \caption{\textbf{Channel-wise amplitude over time.} Absolute amplitude of each channel, median over windows, with shading over the middle half of windows.}
  \label{fig:chan-env}
\end{figure}

\begin{figure}[H]
  \centering
  \includegraphics[width=\linewidth,height=0.85\textheight,keepaspectratio]
    {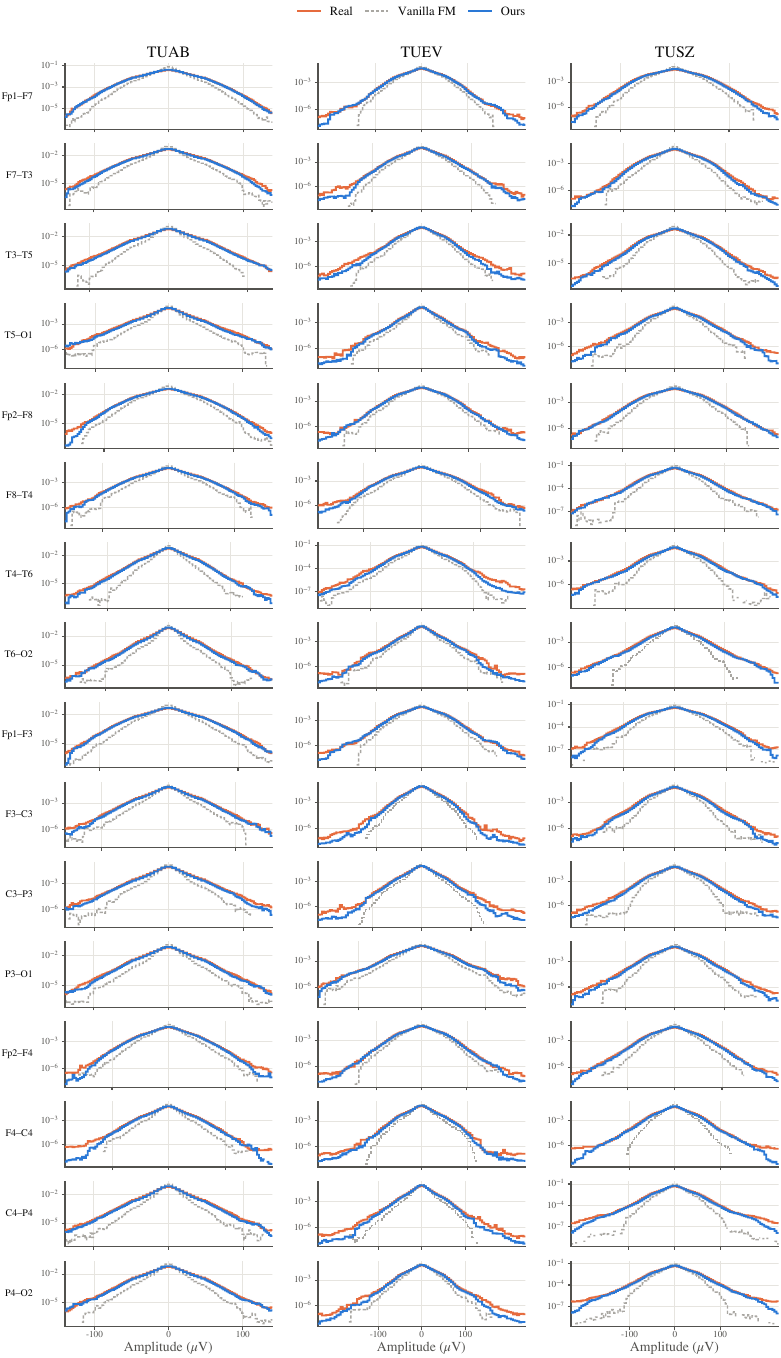}
  \caption{\textbf{Channel-wise amplitude distribution.} Log density of sample amplitudes for each channel.}
  \label{fig:chan-amp}
\end{figure}

Figure~\ref{fig:physio} averages over the sixteen channels. Figures~\ref{fig:chan-psd}, \ref{fig:chan-env} and~\ref{fig:chan-amp} show the same three analyses for each channel separately, so an error on a few channels cannot hide in the average. Each curve is a median over the $2{,}000$ windows of that channel.

\textbf{(1)~Power spectrum:} Figure~\ref{fig:chan-psd} shows the power spectrum of each channel. Real EEG has the most alpha power on P3--O1 and P4--O2 on TUAB and TUEV, and on P3--O1 and Fp1--F7 on TUSZ. \ours{} puts its most alpha power on the same two channels on all three corpora. Vanilla flow matching has less power at almost every frequency, and on TUEV it exaggerates the alpha peak on every channel. Above $12$\,Hz, our spectrum generally falls below the real one but stays within the real shading.

\clearpage
\textbf{(2)~Amplitude over time:} Figure~\ref{fig:chan-env} tracks the amplitude of each channel across the window. Every envelope is flat, for real EEG and for \ours{}, so no channel drifts. Real EEG has the largest amplitude on Fp1--F7, Fp2--F8, Fp1--F3 and Fp2--F4 and smallest on T6--O2, C3--P3 and C4--P4. \ours{} ranks the channels nearly the same way on all three corpora, at $75$--$91\%$ of the real level. Vanilla flow matching stays at $41$--$65\%$.

\clearpage
\textbf{(3)~\mbox{Heavy-tailed} distributions:} Figure~\ref{fig:chan-amp} shows the amplitude distribution of each channel. Our density follows the real curve from the sharp central peak down through the tails on every channel. On TUAB and TUEV, real EEG and \ours{} both have their heaviest tails on Fp1--F7 and Fp2--F8. Above $100\,\mu$V, our samples keep $31$--$86\%$ of the real probability mass per channel, against at most $16\%$ for vanilla flow matching.

\end{document}